\documentclass[10pt,journal,compsoc]{IEEEtran}

\usepackage{times}
\usepackage{epsfig}
\usepackage{graphicx}
\usepackage{amsmath}
\usepackage{amssymb}
\usepackage[ruled]{algorithm2e}
\usepackage{algpseudocode}
\usepackage{caption}
\usepackage{subfigure}
\usepackage{booktabs}
\usepackage{tabularx}
\usepackage{xcolor}
\usepackage{multirow}
\usepackage{multicol}
\usepackage{cite}
\usepackage{hyperref}
\usepackage{bbding}

\begin{document}

\title{TENT: Connect Language Models with IoT Sensors for Zero-Shot Activity Recognition
}

\author{Yunjiao Zhou, Jianfei Yang, Han Zou, Lihua Xie,~\IEEEmembership{Fellow,~IEEE}
        
\thanks{Y. Zhou, J. Yang, Z. Han, and L. Xie are with the School of Electrical and Electronics Engineering, Nanyang Technological University, Singapore (yunjiao001@e.ntu.edu.sg; yang0478@e.ntu.edu.sg; zouh0005@e.ntu.edu.sg; elhxie@ntu.edu.sg).}

\thanks{Jianfei Yang is the project lead and corresponding author.}

}
\markboth{Preprint}%
{Zhou \MakeLowercase{\textit{et al.}}: TENT: Connect Language Models with IoT Sensors for Zero-Shot Activity Recognition}

\maketitle

\begin{abstract}
Recent achievements in language models have showcased their extraordinary capabilities in bridging visual information with semantic language understanding.  
This leads us to a novel question: can language models connect textual semantics with IoT sensory signals to perform recognition tasks, e.g., Human Activity Recognition (HAR)? If so, an intelligent HAR system with human-like cognition can be built, capable of adapting to new environments and unseen categories. This paper explores its feasibility with an innovative approach, IoT-sEnsors-language alignmEnt pre-Training (TENT), which jointly aligns textual embeddings with IoT sensor signals, including camera video, LiDAR, and mmWave. Through the IoT-language contrastive learning, we derive a unified semantic feature space that aligns multi-modal features with language embeddings, so that the IoT data corresponds to specific words that describe the IoT data. To enhance the connection between textual categories and their IoT data, we propose supplementary descriptions and learnable prompts that bring more semantic information into the joint feature space. TENT can not only recognize actions that have been seen but also ``guess'' the unseen action by the closest textual words from the feature space.
We demonstrate TENT achieves state-of-the-art performance on zero-shot HAR tasks using different modalities, improving the best vision-language models by over 12\%.

\end{abstract}


\section{Introduction}
Large language models (LLM) like GPT-3~\cite{brown2020language} and Alpaca~\cite{taori2023alpaca} demonstrate remarkable capabilities in comprehending and processing natural language. Recently, the advancements in language models have contributed significantly to visual sensing tasks, e.g., semantic-guided object recognition~\cite{huang2020pixel, du2022learning} and segmentation~\cite{xu2022simple, yun2023ifseg}, by associating lingual patterns with visual clues.
By the alignment of visual and language representation space using deep neural networks, the visual features are directly mapped into the semantic labels, mimicking the recognition process in a human-like manner. Moreover, the richer semantic language space enables the model to have zero-shot classification capacity~\cite{mishra2020sensors}, i.e., recognizing unobserved objects~\cite{radford2021learning, kojima2022large}.
The amazing progress in computer vision leads us to a compelling question in the Internet of Things (IoT) field: is it possible to connect the sensory signals with language models to attain semantic understandings for IoT tasks? If IoT sensors can be connected to language models, we can force the model to explain what happens behind the IoT data, benefiting various IoT applications, such as smart homes, autonomous robots, and healthcare.


\begin{figure}[t]
	\centering
	\includegraphics[width=1\linewidth]{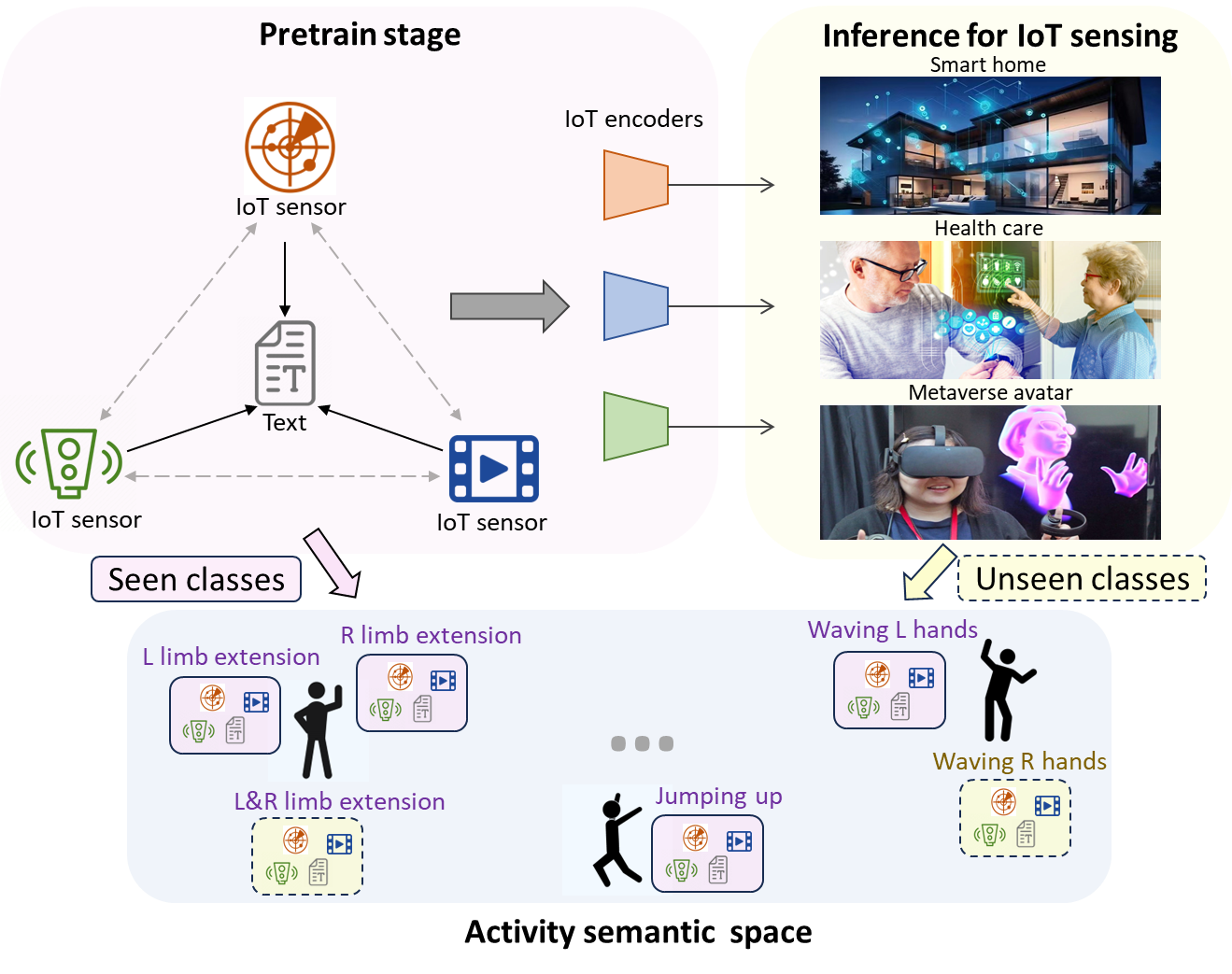}
	\caption{TENT utilizes linguistic supervision to generate multi-modal IoT pre-train models, facilitating the zero-shot applications in various IoT sensing scenarios. The IoT encoders are pretrained with abundant pairs of IoT signals and activity categories, integrating multi-modal information into each uni-modal encoder. These encoders can attend to various IoT sensing tasks separately, even with unseen categories, addressing the limitation of a single modality.}
	\label{fig:illustation}
\end{figure}

To explore the feasibility of connecting language models with IoT sensors, our research is centered on IoT-enabled Human Activity Recognition (HAR) which is a foundational task for various downstream human perception applications~\cite{dang2020sensor, luo2021binarized, zhou2023adapose, zhou2023metafi++}. Current HAR algorithms based on different IoT sensors, such cameras~\cite{jhuang2013towards,zhu2013action}, LiDAR~\cite{roche2021multimodal,luo2020temporal} and RF sensors~\cite{wang2019human, banerjee2020application, luo2023edgeactnet, yang2022securesense, yang2018device, wang2022airfi}, adopt the conventional supervised training using a one-hot vector that represents the category of human activity. Problematically, such a unique numerical identifier that denotes the activity category limits the range of recognizable activities and does not contain any semantic supervision signals. Consequently, existing methods can only recognize human activities that have been trained before, and the prediction is a restrained numerical identifier. 
In this paper, we explore the utilization of textual supervision and aim to align the IoT signal feature space to the semantic language space, which enables zero-shot generalization to unknown classes.
We transform the activity recognition task from the traditional multi-class classification task into a sensor-language similarity matching problem, i.e., retrieving the closest activity category using the IoT sensor data in the semantic space.


However, building a unified feature space between IoT sensory signals and languages encounters several technical challenges. 
Firstly, various IoT sensors are enabled by different physical principles, and thus their data modalities have distinct characteristics. For instance, the camera focuses on colorful appearance; LiDAR delves into spatial voxels; and mmWave radar tracks only dynamic motions, which pose a significant challenge when aligning them with the textual modality. There are some potential solutions, such as IMAGEBIND~\cite{girdhar2023imagebind} that uses images as the intermediary modality to gain indirect alignment between text and other modalities (e.g., depth, thermal data, video), yet it highly relies on high-quality large-scale paired data which is not available in the IoT field.  
Secondly, IoT data is difficult to intuitively understand and interpret. The annotations are more complicated, resulting in fewer sensor-text pairs available compared to vision-language research.
Thirdly, the language models can be heavily influenced by the text prompt during their connection to IoT sensory signals, especially when extending to unseen categories. Crafting the appropriate prompt is crucial to the model adaptation to a spectrum of inputs and tasks.

Confronted with the aforementioned challenges, our approach aims to leverage the rich semantic information inherent in text for HAR while fully harnessing the distinctive attributes of each sensor modality. To align the unique strengths of each modality with the same textual embedding, it is intuitive to establish a unified space that directly bridges the gaps between multi-modal sensor data and text for deep mutual information exchanges. We design a modality-mutual learning strategy that jointly trains multisensor-text pairs to build IoT-semantic space. Through the deep alignment, each uni-modal pre-training encoder becomes more robust by learning from other modalities and forms a unified multi-modal feature space, overcoming limitations in IoT sensing scenarios constrained by a single modality.
Furthermore, we shall address the deficiency of limited textual information in HAR from two perspectives: increasing the volume of textual input and enhancing the model's adaptability to text. These strategies are expected to provide ample and adaptive semantic information, generating more appropriate text prompts for zero-shot HAR.

In this case, we propose an IoT-sEnsors-language alignmEnt pre-Training (TENT) model that utilizes language supervision to enable better HAR of IoT sensors. We offer an effective multi-modal pre-trained method by jointly aligning the sensors of camera, LiDAR, and mmWave with the text embedding into a representative semantic space via contrastive learning, which not only improves the zero-shot performance of each sensor modality but also takes effect in scenarios where not all sensor modalities are available. Additionally, a customized text encoder is designed to capture sensor nuances for activity categories with two key strategies. We embed a structural description explaining the body movement for each class, serving as a complementary prompt to improve the model's authentic comprehension of activity categories. A learnable text embedding is then introduced to generate soft prompts, facilitating the adaptability of the model to different inputs and fostering cross-modal comprehension in depth. Experiments demonstrate the promising feasibility and significant effectiveness of combining language with IoT sensors, proving that language models can be seamlessly extended into the realm of IoT sensors.

The contributions of this work are summarized as follows:
\begin{itemize}
    \item We explore sensor-language learning for IoT sensing scenarios, marking the first work of deeply connecting language models with IoT sensors to enhance cross-modal HAR through text-based semantic information.
    \item We propose a novel method, TENT, that aligns the sensors of RGB, LiDAR, and mmWave simultaneously into a shared textual space using contrastive learning, customized descriptions and soft prompts design. 
    \item Extensive experiments demonstrate that TENT possesses exceptional capabilities in understanding unknown activities and generalizing known ones, achieving remarkable zero-shot performance for HAR.
\end{itemize}
\section{Related work}
\subsection{Cross-modal Language Pre-training Model}
Recent advancements in language models~\cite{brown2020language, taori2023alpaca} have significantly impacted various domains by enhancing natural language understanding and reasoning capabilities. The connection of language models to images, such as CLIP~\cite{radford2021learning} and ALIGN~\cite{jia2021scaling}, has proven that the supervision of language 
exhibits powerful benefits to comprehend and interact with visual representations, enabling open-vocabulary classification~\cite{ghiasi2022scaling, xu2023open}, image-text retrieval~\cite{zhang2020context, chen2020imram} and so on. Building upon the foundation of large-scale pre-trained language models, several recent research fuses language with various visual-related modalities to extend the applicability of language knowledge to a broader spectrum of tasks. For instance, certain works~\cite{ju2022prompting, luo2022clip4clip, xu2021videoclip} have aligned semantic embeddings with spatial-temporal visual features, enabling tasks like video retrieval and classification through the utilization of text. Simultaneously, other research~\cite{zhang2022pointclip, hong20233d} has extended language models into 3D world. They achieve this by mapping 3D point clouds into multi-view 2D image features, indirectly aligning them with the help of image-text pre-training guidance.
Additionally, IMAGEBIND~\cite{girdhar2023imagebind} harnesses images as an intermediary modality, enabling alignment between text and a multitude of multi-modal visual-related modalities, including audio, depth, thermal, and Egocentric Videos. This opens avenues for cross-modal retrieval and reasoning across multiple modalities.

However, a critical limitation of current methods lies in their heavy reliance on the effectiveness of aligning images with text. Particularly in tasks that demand fine-grained IoT perception, where images might struggle to capture comprehensive spatial information, these approaches often fail to achieve reliable alignment with language. Instead, TENT takes an approach to directly align IoT sensors with language, aiming to derive informative representations from multiple sensors while mitigating the limitations associated with image-based alignment.

\subsection{Zero-shot Human Activity Recognition}
Zero-shot HAR~\cite{chen2022salience} is a crucial area in IoT applications. While existing studies~\cite{li2010action,yang2014effective,yang2022autofi,yang2022efficientfi} have achieved precise activity estimation in known categories, traditional supervised classification methods have limitations when dealing with unseen activity classes, hindering their broader application in the real world. To enhance zero-shot capabilities, current efforts~\cite{lampert2009learning,xu2020attribute, mandal2019out, chen2021elaborative} mainly adopt semantic space-based approaches by introducing category information for the knowledge transfer to unseen classification. 
~\cite{lampert2009learning} firstly tackles the zero-shot problem with manual-defined attributes, which provides high-level characters to transfer the knowledge from seen to unseen classes. Nonetheless, it heavily relies on subjective human judgment and can involve significant labor to identify suitable attributes.
Recently, many works~\cite{mandal2019out, lin2022cross, qin2017zero} leverage word embeddings of category names to provide supplementary information. For instance, ~\cite{mandal2019out} employs adversarial mechanisms to determine whether features belong to unseen classes, enabling concise and effective zero-shot classification without requiring extensive manual processing. However, the brief activity category may offer vague and misleading guidance, posing a significant challenge for fine-grained and highly variable activities. In contrast, our work aims to generate robust and representative semantic embeddings that vividly capture the spatial distribution of activity categories, fostering the widespread of activity sensing.

\section{Method}\label{sec:method}
\begin{figure*}[t]
	\centering
	\includegraphics[width=0.8\linewidth]{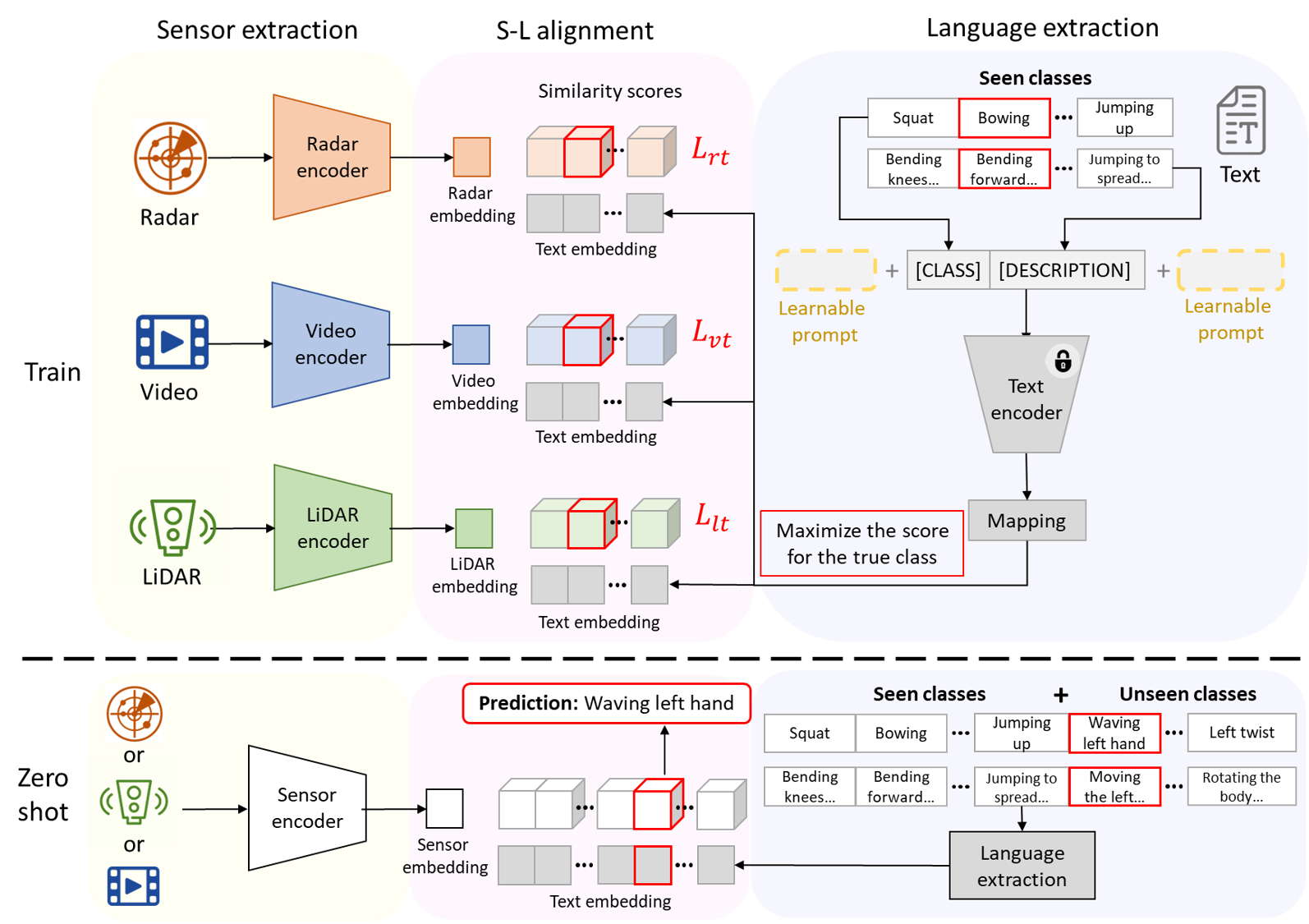}
	\caption{Overview of IoT-sEnsors-language alignmEnt pre-Training (TENT), which consists of three modules of sensor embedding extraction, language embedding extraction and Sensor-Language alignment. Distinct feature extractors are tailored for different sensors, while text, tokenized customarilly via complementary description and learnable prompt, is extracted by a frozen encoder and a mapping layer. During the training stage, TENT calculates the similarity scores between embedding of three sensors and text embedding, and maximizes the similarity scores for the ground-truth class. At test time TENT can make zero-shot predictions for any sensor modality based on the highest similarity scores.}
	\label{fig:framework}
\end{figure*}

TENT introduces a novel multi-sensor pre-training pipeline that leverages linguistic supervision for HAR. As shown in Fig.~\ref{fig:framework}, TENT is structured into three fundamental components: sensor embedding extraction, language embedding extraction, and sensor-language alignment. Sensor embedding extraction captures distinctive features from each modality, while language embedding extraction generates adaptive prompts reflecting specific activity information for each category. Then sensor-language alignment serves as a bridge to connect sensor and language representations in a shared space, realizing cross-modal knowledge transferring. In the training phase, a joint training approach aligns video, LiDAR, and mmWave sensor modalities with text through contrastive loss to gain pre-train encoders for each sensor. During testing, inference for unseen categories can be conducted independently using the encoder specific to each modality, avoiding the necessity of acquiring all the data modalities. Such pre-training strategy integrates useful information from different aspects of sensors to each single modality, playing an important role in various IoT application scenarios with only one sensor modality. In the subsequent sections, we will present detailed descriptions of each component.

\subsection{Sensor Embeddings Extraction}
Given an IoT activity recognition dataset consisting of multiple signal streams of IoT sensors, we divide it into seen set ($\mathcal{D}$) with $k_s$ training classes and unseen set ($\mathcal{\hat{D}}$) from $k_u$ disjoint label classes. It is important to note that these two sets have no overlapping categories, indicated by the constraint $\mathcal{D} \cap \mathcal{\hat{D}} = \emptyset$. This separation represents a zero-shot learning scenario, where the objective is to recognize activities not seen during the training phase.

For the seen set ($\mathcal{D}$), the data input is structured as tuples of all the sensor modalities and their corresponding activity categories $I^t_i$ over a period of time series $S$, including video $I^v_i \in \mathbb{R}^{3\times224\times224\times S}$, LiDAR $I^l_i \in \mathbb{R}^{3\times N_l\times S}$ and mmWave radar $I^r_i \in \mathbb{R}^{5\times N_r\times S}$, denoted as $\mathcal{I}_i = (I^v_i, I^l_i, I^r_i, I^t_i)$, where $N_l$ and $N_r$ represent the variable number of point clouds in LiDAR and mmWave radar. While the unseen set ($\mathcal{\hat{D}}$) contains a flexible number of the available sensor modalities to represent various application scenarios. Thus the input of the unseen set can be written as $\mathcal{\hat{I}}_i = (\hat{I}^m_i, \hat{I}^t_i), m\in(v,l,r)$, where $m$ is the dynamic subset of all the sensors. TENT aims to predict the correct activity categories based on $\mathcal{\hat{I}}_i$ from the unseen set.

Due to the distinct data structures and characteristics of each modality, sensor embedding extraction module is designed with separate encoder for each modality. These encoders map different sensor inputs to the common-sized embeddings $E^m_i \in \mathbb{R}^{768}, m \in (v, l, r)$, effectively extracting modality-specific activity information. To accomplish this, we leverage established frameworks for each modality as the backbone for our sensor encoders ($\mathcal{F}$). Specifically, we employ TimesFormer~\cite{bertasius2021space} for video data, harnessing its attention mechanisms to efficiently capture spatio-temporal patterns of videos. Meanwhile, for LiDAR and mmWave radar data, which present unique characteristics in the form of point clouds, Point Transformer~\cite{zhao2021point} is adopted to model the complex spatial dependencies within 3D data by point-wise interaction between neighbors via self-attention mechanisms. It has strong contextual understanding, demonstrating state-of-the-art performance in various point cloud recognition tasks. Thus the process of sensor embedding extraction can be expressed as:
\begin{equation}
    E^m_i = \mathcal{F}^m (I^m_i), \quad m \in (v,l,r).
\end{equation}
The sensor embedding extraction captures the representative features of the IoT sensors, which contain descriptive temporal and spatial information based on different sensor perspectives. It forms a crucial foundation for the subsequent cross-modal alignment.

\subsection{Language Embedding Extraction}
Language embedding extraction is proposed to generate semantically meaningful and robust text embeddings for both seen and unseen classes, aiming to reflect the spatial distribution of activity categories based on the semantic distance. To achieve this, we leverage the pre-trained text encoder ($\mathcal{F}_t$) sourced from CLIP~\cite{radford2021learning} as our backbone, transferring its powerful ability of natural language understanding to IoT sensing. Trained on extensive textual data, it is able to capture diverse semantic information effectively without further modification.  

Our dataset ($\tilde{C}_k=C \cup \hat{C}, k \in \{1, 2,..., k_s+k_u=27\}$) includes 22 activity categories used for training ($C_{a}, a \in \{1, 2,..., k_s=22\}$), with an additional 5 unseen categories introduced for zero-shot testing ($\hat{C}_{b}, b \in \{1, 2,..., k_u=5\}$). Consequently, the available textual semantic information is relatively sparse and limited, where the model can hardly gain enough prior knowledge of the classes. To address it, we introduce descriptions ($\tilde{D}_k=D \cup \hat{D}, k \in \{1, 2,..., k_s+k_u=27\}$) for each activity category to provide supplementary information, enriching the encoder's semantic understanding of activity categories. For instance, ``Bowing'' is described as ``an activity of a person bending forward at the waist". These descriptions offer detailed explanations of the specific body movements involved in each activity, providing a more reliable basis for constructing the embedding spatial distribution of activity categories. The names together with descriptions of each class are fed into a tokenizer to generate the text token $t_a$ for further extraction:
\begin{equation}
    t_{a} =\oplus(\phi(C_{a}),\phi(D_{a})),
\end{equation}
where $\phi$ represents the tokenizer to convert text into numerical representations for computational processing, and $\oplus$ is the concatenation operation.

However, the hard combination between the categories and descriptions may lead to an overfitting issue towards the seen set, decreasing the zero-shot performance on the unseen set. Hence, motivated by CoOp~\cite{zhou2022learning}, we introduce learnable prompts to dynamically generate prompts that benefit the understanding of textual information in a flexible manner, enhancing the model's adaptability and robustness for unseen text. In TENT, $n$ learnable prompts are placed both before and after the text token $t_a$ to adapt to the specific context of the data, ensuring that the model can effectively leverage textual cues during activity recognition tasks. The token of each category $T_a$ is gained via the concatenation of learnable prompt $p_n$ and text token $t_a$:
\begin{equation}
    T_a = \oplus(p_1,,,p_n, t_a, p_{n+1},,,p_{2n}),
\end{equation}
where $p_n$ is the $n$-th learnable prompt in the tokens, and $n$ is set to be 16 in our experiments.

To extract the representative text embeddings, the tokens are firstly extracted by the fixed text encoder ($\mathcal{F}_t$), which can understand vast numbers of natural language to produce the semantic embeddings. As the objective of TENT is to align the language model to IoT sensors, the embeddings are supposed to go through a mapping layer ($\mathcal{G}$) to fine-tune the text embeddings, driving the semantic embedding to IoT sensor space.
Therefore, the formula of language embedding extraction is denoted as:
\begin{equation}
    E^t_a = \mathcal{G}(\mathcal{F}_t(T_a)).
\end{equation}
Language embedding extraction harnesses the text processing capabilities of large language models to effectively generate embeddings from textual activity categories. This process exhibits strong inference efficacy, even for unseen categories, enabling robust reasoning. It can effectively map the unseen classes into the semantic space shared with seen ones based on semantic similarity, greatly enhancing the model's ability to comprehend activities accurately within IoT sensing applications.

\subsection{Modality-Mutual Learning}
After obtaining feature embeddings of both sensor modalities and language, we propose modality-mutual learning to realize sensor-language alignment, bridging the gap between languages and different sensor modalities. Specifically, we design a joint optimization approach that refines the parameters in both language extraction and sensor encoders for the three modalities. Such an optimization process follows a contrastive strategy, wherein the objective is to encourage sensor embeddings to closely align with their respective text embeddings for the same activity class while maintaining separation from unrelated classes.
In consequence, the text embedding will be mapped into the IoT sensor domain, comprehensively capturing the distinctive characteristics of all sensor modalities. 
To achieve these goals, we calculate similarity scores between text embeddings and three sensor modalities. This is targeted to maximize the similarity between the embeddings from the same activity category while minimizing the similarity between embeddings from different categories. The process is realized via infoNCE~\cite{oord2018representation} losses $\mathcal{L}^{mt}$ for each sensor modality:
\begin{equation}
    \small
\mathcal{L}^{mt}=-\log \frac{\exp(E^m_i 
\cdot \{E^t_a\}^+/\tau)}{\sum_{E^t_a \in \{\{E^t_a\}^+, \{E^t_a\}^-\}}\exp(E^m_i \cdot E^t_a/\tau)},
\end{equation}
where $m \in (v,l,r)$, $\tau$ refers to a scalar temperature, $\exp()$ is the exponential function, and $\{E^t_a\}^+, \{E^t_a\}^-\}$ denote the positive and negative text embeddings overlapping with sensor embeddings $E^m_i$ respectively. The term $\exp(E^m_i \cdot \{E^t_a/\}^+\tau)$ measures the similarity between the embedding of a specific sensor modality and the corresponding text embedding, while the denominator considers similarities with all the categories. By minimizing this loss, we ensure that the model assigns high similarity scores to the correct category pairs. 

The training objective of TENT involves integrating the contrastive losses of each sensor modality simultaneously:
\begin{equation}
\mathcal{L}=\alpha\mathcal{L}^{vt}+\beta\mathcal{L}^{lt}+\gamma\mathcal{L}^{rt},
\end{equation}
where $\alpha$, $\beta$ and $\gamma$ are the hyper-parameters that assign different weights for the sensor modalities. These weights determine the balance among the three sensor modalities during the alignment process, essentially indicating which modality the text embedding should pay more attention to. In our practical experiments, we observed that LiDAR and Radar contain much 3D information, making them more robust in zero-shot scenarios. Consequently, we assign higher weights to these two modalities.

In summary, sensor-language alignment effectively bridges the gaps between sensor data and linguistic semantics through joint training, facilitating the model's cross-modal comprehension of activity recognition. It drives the modality information to exchange with each other, integrating the multi-modal and linguistic perspective into each single sensor encoder. Therefore, when a single modality encoder is used for inferring unseen activity categories, it can leverage the perspectives from all sensor modalities for comprehensive reasoning, thus achieving a more robust zero-shot performance.

\subsection{Zero-Shot Human Activity Recognition}
Building on the robust multi-modal pre-training model through joint training, TENT's primary goal is to produce general encoders for each sensor modality which can predict the correct results for unseen activity categories. Unlike conventional methods that usually require extensive retraining or demand labeled data for new activity classes, TENT tackles these tasks without further training process, contributing to great convenience and widespread application of IoT sensing.

During testing, it is likely that not all the sensor modalities are accessible. Taking an unseen class of LiDAR signal ($\hat{I}_i^l$) for example, all the parameters in the test are fixed and adopted with the pre-training modal. TENT first exacts robust LiDAR embedding ($\hat{E}_i^l$) via the pretrained LiDAR encoder ($\mathcal{F}^l$), which is integrated with multi-modal information from the other two sensors with the help of the alignment to IoT sensor semantic space:
\begin{equation}
    \hat{E}^l_i = \mathcal{F}^l(\hat{I}^l_i).
\end{equation}
As for language embedding extraction, we feed both seen and unseen classes into the encoder, aiming to achieve powerful activity recognition on both seen and unseen categories. The text embedding in the zero-shot stage is generated by:

\begin{gather}
        \hat{t}_k = \oplus(\phi(\tilde{C}_k), \phi(\tilde{C_k})),\\
        \hat{T}_k = \oplus(p_1,,,p_n, \hat{t}_k, p_{n+1},,,p_{2n}),\\
        \hat{E}_k^t = \mathcal{G}(\mathcal{F}_t(\hat{T}_k)).
\end{gather}

Finally, through similarity calculation between LiDAR embedding $\hat{E}^l_i$ and text embeddings of all the categories $\hat{E}^t_k$, TENT makes the activity prediction $\hat{y}$ based on the highest similarity score among the $k$ activity categories.
\begin{equation}
    \hat{y} = \arg\max_k (\hat{E}^l_i \cdot \hat{E}^t_k).
\end{equation}
In a nutshell, these simple procedures ensure TENT can be applied in a wide range of uni-modal and multi-modal scenarios, where each sensor encoder predicts accurate activities from the unified multi-modal feature space, thereby promoting the further adoption and development of IoT sensing. 
\section{Experiment}
\subsection{Setup}
\begin{figure}[t]
	\centering
	\includegraphics[width=1\linewidth]{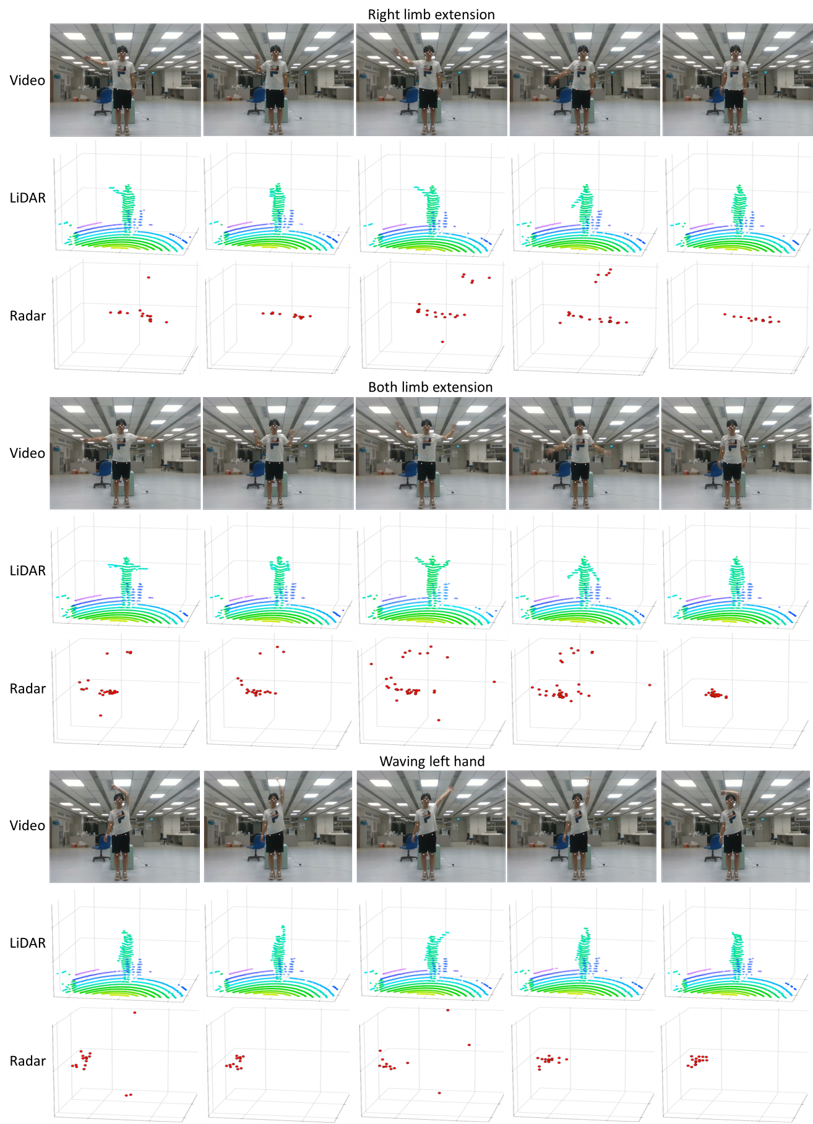}
	\caption{The visualization of three activities in MM-Fi. Two limb extension actions are similar and waving left hand is quite different, which is reflected by the latent space in Figure~\ref{fig:tsne}.}
	\label{fig:dataset}
\end{figure}

\textbf{Dataset.} To enable multi-sensor alignment, there is a high requirement for a dataset with a substantial number of sensor pairs and synchronized IoT signals. We adopt the recent multi-model non-intrusive human dataset denoted MM-Fi~\cite{yang2023mm}, which offers much convenience for the connection between language and IoT sensors, to evaluate the performance of TENT. As depicted in Fig.~\ref{fig:dataset}, it comprises various IoT signals, including images, LiDAR, and mmWave radar point clouds, of 27 daily or rehabilitation activity categories. MM-Fi invites 40 subjects to perform each of the 27 activities for 30 seconds at 4 different environmental scenes, serving as a large public human dataset with 320k synchronized frames for comprehensive evaluations. In our experiments, we aggregate data from each sensor modality by combining 8 consecutive frames to reflect the temporal information as each modality input, contributing to totally 8640 sequences for each modality. 

To enable a robust zero-shot evaluation, 5 classes, namely Left twist (Z01), Both limb extension (Z02), Right side lunge (Z03), Waving left hand (Z04), and Right side throwing (Z05), are excluded from the training set and only used for testing. This selection aligns with the core principle of zero-shot learning: ensuring unknown categories differ from known ones but share certain similarities. For example, the Right twist in a known category and the Left twist in an unknown category both rotate the body, but the direction of the rotation is different and symmetrical to each other. During the zero-shot interference stage, the network can classify unknown classes reasonably based on their proximity to the known class label space.

\textbf{Experimental details.} 
In the experimental setup, each image frame is resized to a standard 224x224 pixel size to meet the input requirements of TimesFormer. To manage varying numbers of point clouds in LiDAR and radar clips, we balanced the number of point clouds within each batch based on the maximum point cloud count, ensuring a consistent data structure. Additionally, the loss weights for video, LiDAR, and radar are set at 0.4, 1.3, and 1.3, respectively, ensuring the unified feature space containing many 3D spatial details. 

Our experiments were conducted using the PyTorch framework, with optimization performed through the Stochastic Gradient Descent with Momentum (SGDM) algorithm. Key hyperparameters include a batch size of 10, a learning rate of 0.001, a momentum value of 0.9, and a weight decay of 0.0005.
The training process spanned 50 epochs, and we implemented a step-wise learning rate schedule with reductions occurring at the 20th, 30th, and 40th epochs.

\subsection{Overall Performance}

\begin{table*}[htbp]
\center

\caption{Top 1 accuracy comparison for zero-shot performance on MM-Fi dataset.}
\label{table:backbone}
\scalebox{1}{
\begin{tabular}{c|c|c|cccccc}
\toprule
\multirow{2}{*}{\textbf{Modality}} & \multirow{2}{*}{\textbf{Method}} & \multirow{2}{*}{\textbf{Seen set}} & \multicolumn{6}{c}{\textbf{Unseen set}}        \\
                          &     &                             & \textbf{Z01} & \textbf{Z02} & \textbf{Z03} & \textbf{Z04} & \textbf{Z05} & \textbf{Avg}\\
\midrule
 \multirow{3}{*}{Video} & Vanilla                                 &       88.8                                           & 0.0       & 0.0   & 0.0   & 0.0   & 0.0   & 0.0   \\    
 & CEWGAN-OD~\cite{mandal2019out} & 88.9 & 1.3 & \textbf{12.1} & 9.7 & 9.6 & 3.3 & 7.25\\
  & I-VL~\cite{ju2022prompting}                                 &     59.6                                              &  2.8 &  0.0   &   6.6  &  32.8   &  6.3    &    8.5    \\
 & CLIP4Clip~\cite{luo2022clip4clip}                                 & 89.7 & 0.9 & 2.5 & 4.7 & 76.9 & 8.7  & 18.4\\

  & CoOp~\cite{zhou2022learning}             &         90.3                                                    &    10.3 & 2.8 & 2.5 & \textbf{ 77.5} & 2.2       &     19.8   \\
 & \textbf{TENT}   &  88.2   &  \textbf{10.6 }  &  7.2   &  \textbf{42.2}   &  73.1   &  \textbf{ 23.4 }  &      \textbf{32.4} \\
\midrule
\multirow{3}{*}{Lidar}   & Vanilla                                 &       90.5                                           & 0.0       & 0.0   & 0.0   & 0.0   & 0.0   & 0.0   \\
& CEWGAN-OD* & 91.3 & 0.3 & 46.3 & 15.6 & 18.8 & 5.3 & 17.6 \\
& CLIP4Clip*                                & 92.2 & 1.0 & 44.8 & 39.9 & 36.4 & 5.8  & 26.5\\
& CoOp*                                 &           91.9                                      &    4.4     &   0.9  &  38.1   &   31.6  &  0.6   &   15.4   \\
 & \textbf{TENT} & 92.8     &   \textbf{14.1}  &   \textbf{73.4}  &  \textbf{51.6}   &  \textbf{76.6}   &  \textbf{36.6}  &     \textbf{50.0}  \\
\midrule
\multirow{3}{*}{Radar}   & Vanilla                                 &       81.7                                           & 0.0       & 0.0   & 0.0   & 0.0   & 0.0   & 0.0   \\
& CEWGAN-OD* & 87.6 & 0.8 & 28.8 & 9.7 & 16.8  & 2.4 & 11.7 \\
& CLIP4Clip*                                 & 87.6 & 10.4 & \textbf{51.0} & 32.8 & 50.6 & 6.5  & 31.8\\
& CoOp*                                 &        88.4                                          &  \textbf{ 23.4}     &  3.1   &   44.4  &  47.5   &   11.6  &   26.0  \\
 & \textbf{TENT}   &     88.1     &   21.9  &  47.5   &   \textbf{47.8}  &   \textbf{71.3}  &  \textbf{15.6}   &      \textbf{43.9} \\
\bottomrule
\end{tabular}}
\end{table*}
To assess the necessity of connecting the language model to IoT sensors, we first conduct preliminary experiments to evaluate the individual performance of each modality's backbone denoted as Vanilla. These experiments follow the conventional approach of training each modality separately for one-hot classification. As shown in Table~\ref{table:backbone}, the vanilla backbones in all three modalities demonstrate excellent recognition performance within their training knowledge. However, they lose the effect when confronted with entirely unfamiliar activity categories. This limitation is a common challenge in one-hot classification, which arises from the fact that these models lack any prior information about unknown categories, making zero-shot categorization a difficult task.

Recent visual-based HAR works prove that great improvement in zero-shot can be achieved by the introduction of language embeddings. In this manner, we make extensive comparisons between the current SOTA methods and TENT. Due to the absence of dedicated zero-shot algorithms for Lidar and mmWave in activity recognition, we replicated algorithms from the video domain for comparison and marked them with an asterisk ($*$). As observed in table~\ref{table:backbone}, both the vanilla model and zero-shot algorithms achieve competitive and high accuracy of activity recognition on the seen set.

In contrast, it is crucial to note that only the language-related methods enable zero-shot ability for the unseen activity categories. This underscores the importance of incorporating language information to construct a semantic space that accurately reflects the distribution of activity categories. Among them, TENT brings a significant enhancement to the average of zero-shot performance across all the modalities, improving the average results of video, LiDAR and Radar by exceeding 12\%, 24\% and 12\%, respectively. The great increase may stem from the joint pre-training strategy that enables each uni-modal encoder to exact the feature embeddings from the unified multi-modal perspectives, surmounting the constraints of the specific modality. The above findings demonstrate that language models can effectively connect various IoT sensors to a joint feature space, transferring the semantic information of activity categories to the sensor modalities for unseen classes.

\subsection{Cross-Domain Evaluation}
\begin{table*}[htbp]
\center

\caption{The performance of TENT with different dataset split settings.}
\label{table:cross}
\scalebox{1}{
\begin{tabular}{c|c|c|cccccc}
\toprule
\multirow{2}{*}{\textbf{Setting}} & \multirow{2}{*}{\textbf{Modality}} & \multirow{2}{*}{\textbf{Seen set}} & \multicolumn{6}{c}{\textbf{Unseen set}}        \\
                          &     &                             & \textbf{Z01} & \textbf{Z02} & \textbf{Z03} & \textbf{Z04} & \textbf{Z05} & \textbf{Avg}\\
\midrule
 \multirow{3}{*}{Random Split} & Video                                 & 88.2 & 10.6 & 7.2 & 42.2 & 73.1 & 23.4  & 32.4\\    
 & LiDAR & 92.8 & 14.1 & 73.4 & 51.6 & 76.6 & 36.6  & 50.0\\
 & Radar      & 88.1 & 21.9 & 47.5 & 47.8 & 71.3 & 15.6 & 43.9\\

\midrule
 \multirow{3}{*}{Cross-Subject Split} & Video                                 &       91.0                                                  & 12.5   & 3.1   & 12.5   & 48.4   & 50.0  & 27.5 \\    
 & LiDAR & 93.2  & 3.1 & 68.8 & 56.2 & 73.4 & 46.9 & 49.1\\
 & Radar      &        86.0                  & 34.4 & 34.4 & 50.0 & 64.1 & 20.3  & 43.1\\
\midrule
 \multirow{3}{*}{Cross-Environment Split} & Video                                 &       57.8                                                 & 6.2   & 16.2   & 17.5   & 25.0   & 25.0  & 17.5 \\    
 & LiDAR & 65.7 & 28.7 & 42.5 & 72.5 & 21.3 & 42.5  & 40.7\\
 & Radar                                & 86.8 & 32.5 & 20.0 & 20.0 & 52.5  & 15.0  & 29.5\\
\bottomrule
\end{tabular}}
\end{table*}
\begin{table*}[t]
\center

\caption{Ablation results for TENT on MM-Fi Dataset.}
\label{table:ablation}
\begin{tabular}{c|ccc|c|c|cccccc}
\toprule
& \multirow{2}{*}{\textbf{Joint}} & \multirow{2}{*}{\textbf{Description}} & \multirow{2}{*}{\textbf{Soft prompt}} & \multirow{2}{*}{\textbf{Modality}} & \multirow{2}{*}{\textbf{Seen set}} & \multicolumn{6}{c}{\textbf{Unseen set}} \\
                       & &     & &                      &                               & \textbf{Z01} & \textbf{Z02} & \textbf{Z03} & \textbf{Z04} & \textbf{Z05} & \textbf{Avg} \\
\midrule

\multirow{3}{*}{1} & \multirow{3}{*}{\XSolidBrush} & \multirow{3}{*}{ \XSolidBrush} & \multirow{3}{*}{ \XSolidBrush} &  Video & 89.7 & 0.9 & 2.5 & 4.7 & 76.9 & 8.7  & 18.4\\
                & & & & LiDAR & 92.2  & 1.0 & 44.8 & 39.9 & 36.4 & 5.8 & 26.5\\
                & & & & Radar & 87.6  & 10.4 & 51.0 & 32.8 & 50.6 & 6.5 & 31.8\\
                  \midrule
\multirow{3}{*}{2} & \multirow{3}{*}{\XSolidBrush} & \multirow{3}{*}{ \Checkmark} & \multirow{3}{*}{ \XSolidBrush} &  Video & 89.9  & 2.8 & 1.3 & 16.2 & 77.5 & 36.6 & 25.5\\
                & & & & LiDAR & 92.8  & 5.2 & 20.1 & 55.8 & 52.3 & 34.1 & 34.8\\
                & & & & Radar & 88.0  & 15.3 & 53.2 & 42.2 & 56.8 & \textbf{19.5} & 38.5\\
                  \midrule
\multirow{3}{*}{3} & \multirow{3}{*}{\XSolidBrush} & \multirow{3}{*}{ \Checkmark} & \multirow{3}{*}{ \Checkmark} & Video & 89.3  & 4.4 & 1.3 & 25.6 & 78.1 & 20.0 & 25.1\\
                 & & & & LiDAR & 93.1  & 7.5 & 19.1 & \textbf{57.5} & 58.1 & 14.7 & 31.2\\
                 & & & & Radar & 88  & 18.1 & 34.1 & 36.6 & 50.0 & 11.6 & 32.5\\
                  \midrule
\multirow{3}{*}{4} & \multirow{3}{*}{\Checkmark} & \multirow{3}{*}{ \XSolidBrush} & \multirow{3}{*}{ \XSolidBrush} & Video & 89.9 & 0.6 & 4.1 & 0.9 & 79.4 & 13.1 & 18.8\\
                 & & & & LiDAR & 93.5 & 7.5 & 57.5 & 53.8 & 59.1 & 22.5 & 38.2\\
                 & & & & Radar & 87.7 & 19.1 & 45.6 & 29.7 & 56.9 & 6.6 & 34.3\\
                  \midrule
\multirow{3}{*}{5} & \multirow{3}{*}{\Checkmark} & \multirow{3}{*}{ \Checkmark} & \multirow{3}{*}{ \XSolidBrush} & Video & 90.5 & 1.3 & 1.3 & 6.6 & 71.9 & \textbf{48.1} & 26.1\\
                 & & & & LiDAR & 92.8 & \textbf{22.5} & 43.4 & 55.0 & 57.2 & \textbf{41.6} & 40.8\\
                 & & & & Radar & 87.7 & 10.3 & \textbf{61.6} & 40.9 & 59.1 & 21.9 & 41.3\\
                  \midrule
\multirow{3}{*}{6} & \multirow{3}{*}{\Checkmark} & \multirow{3}{*}{ \XSolidBrush} & \multirow{3}{*}{ \Checkmark} & Video & 89.2 & \textbf{11.6} & \textbf{7.8} & 2.5 & \textbf{85.3} & 14.4 & 22.1\\
                 & & & & LiDAR & 91.1 & 2.8 & \textbf{75.0} & 49.1 & 70.6 & 21.6 & 44.4\\
                 & & & & Radar & 87.7 & 13.8 & 30.3 & 12.8 & 38.1 & 6.2 & 23.5\\
                  \midrule
\multirow{3}{*}{TENT} & \multirow{3}{*}{\Checkmark} & \multirow{3}{*}{ \Checkmark} & \multirow{3}{*}{ \Checkmark} &  Video & 88.2 & 10.6 & 7.2 & \textbf{42.2} & 73.1 & 23.4  & \textbf{32.4}\\
               & & &  & LiDAR & 92.8 & 14.1 & 73.4 & 51.6 & \textbf{76.6} & 36.6  & \textbf{50.0}\\
               &  & & & Radar & 88.1 & \textbf{21.9} & 47.5 & \textbf{47.8} & \textbf{71.3} & 15.6 & \textbf{43.9}\\
                  \bottomrule
\end{tabular}
\end{table*}

Our experiments are conducted under the random split setting by default, where the training and testing sets share the same data distribution, including subjects and environments. However, in practical applications, new subjects or experimental contexts may often emerge in the testing set, posing challenges for zero-shot HAR. In such scenarios, we investigate TENT's enhanced zero-shot capability under more complex conditions, with a specific focus on its recognition of unknown activities for new subjects or experimental scenes. MM-Fi dataset offers three data split strategies, consisting of Random Split (80\% for training and 20\% for testing), Cross-Subject Split (32 subjects for training and 8 for testing) and Cross-Environment Split (3 environments for training and 1 for testing). Table.~\ref{table:cross} shows the performance of TENT on the seen and unseen sets with the different split settings. 

Among these settings, TENT exhibits the best zero-shot performance under the Random Split setting across all three modalities. This is attributed to the prior exposure of the network to the characteristics of the subjects and environments present in the testing set. In the Cross-Subject Split setting, where new subjects are introduced and the environmental context remains consistent with the training data, TENT achieves high zero-shot results similar to those in the Random Split setting. While the Cross-Environment Split setting, where the testing data features new environmental contexts but with subjects seen during training, suffers a consistent decline in performance, the zero-shot scores still surpass those of the current state-of-the-art methods shown in Table.~\ref{table:backbone}.
In a nutshell, these results highlight TENT's robustness and adaptability in zero-shot performance, even in challenging scenarios involving new subjects or environments. Such advancement extends the applicability of HAR in real-world IoT sensing, improving the effective recognition of new activity categories in new scenarios with an increasing number of additional subjects.

\subsection{Ablation Study}

To verify the effectiveness of TENT, extensive ablation experiments are conducted on the trained and unseen set of MM-Fi dataset with results demonstrated in Table~\ref{table:ablation}.
All experiments are conducted with linguistic supervision, where "joint" refers to whether the three modalities are jointly trained, and "description" and "soft prompt" denote customized language extraction components discussed in Section~\ref{sec:method}. The bold figures represent the best results for each modality in the specific unseen categories. The influence of joint training is significant. Models trained with a joint approach consistently outperform those trained separately on individual modalities with the same setting of language extraction. It is evident that joint training enables the model to create a more expressive and information-dense semantic space by integrating video, LiDAR, and radar modalities. Consequently, all three modalities exhibit notable performance enhancements in terms of zero-shot accuracy.
Secondly, as for the design choices within the language extractor, the introduction of textual descriptions yields a remarkable boost in zero-shot performance across all modalities. For instance, compared to the same experimental setting, descriptions improve the overall zero-shot accuracy of mmWave radar by a large margin of exceeding 7\%. This enhancement arises from the richer semantic information embedded in descriptions, which allows for a clearer representation of semantic affinities among activity categories. Thus, cross-modal alignment is driven closer to the true distribution of IoT scenes, enhancing the model's adaptability to new classes. Additionally, the incorporation of soft prompts empowers the model to adaptively generate text embeddings. Though it may cause instability when used independently, soft prompt exhibits a significant increase in performance when combined with joint training and descriptions. With rich semantic and modality information, soft prompt effectively balances the relationships among various modalities, aligning the text semantic space closely with the embeddings of IoT sensors, thus enabling the model to achieve more stable and robust zero-shot recognition results. 
The ablation study proves the effectiveness of TENT with joint training across modalities and the customized designs of the language extractor. These results emphasize the necessity of establishing a shared semantic space to guide multi-modal recognition and highlight the significance of detailed descriptions and soft prompts in amplifying zero-shot performance.

\subsection{Visualization Analysis}
\begin{figure}[htbp]
	\centering
	\includegraphics[width=0.8\linewidth]{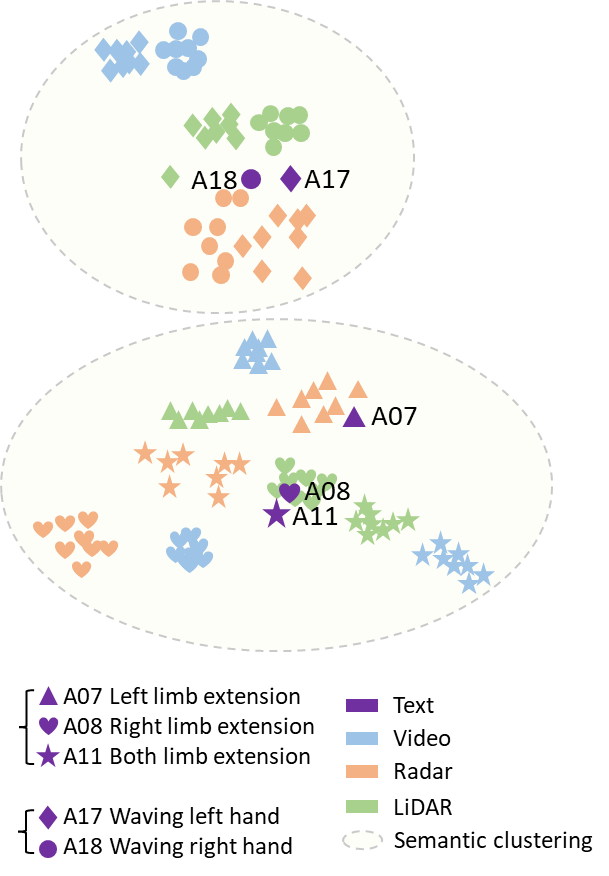}
	\caption{T-SNE visualization of the joint feature space.}
	\label{fig:tsne}
\end{figure}
To better illustrate TENT's capacity to effectively encode textual information and establish a Sensor-Language unified space, we conducted a visualization analysis through mapping the embeddings from various modalities of selected categories into a shared space, as shown in Figure~\ref{fig:tsne}.
It is evident that the distributions of IoT sensor embeddings cluster near their corresponding textual features within the same category, which proves TENT's effectiveness in bridging the gaps between semantic information and the IoT sensing domain.
Furthermore, the visualizations offer a clear depiction of the semantic space distances between different activity categories. For instance, activities that share similarities, but differ mainly in direction, tend to exhibit relatively close spatial distributions. Conversely, activities that are notably distinct are often situated far apart from other activities. Notably, for unseen classes like A11 and A17, TENT successfully maps them into the correct clusters of limb extension and waving hands, respectively, based on semantic information. This illustrates the accuracy of TENT to position activity categories in the spatial context, achieving robust zero-shot alignment. 
\section{Conclusion}
In this work, we propose a multi-modal pre-training model that first proves the feasibility and effectiveness of the language model to connect IoT sensors for zero-shot activity recognition. We design a joint training pipeline through directly aligning text embedding with a series of IoT sensors, leading to a unified semantic space with instructive representations learned from multiple modalities. To enhance the spatial mapping of semantic features, customized language extraction is proposed with a supplementary description to provide informative details of categories and a learnable prompt to ensure the robustness of semantic embedding. 
Our extensive experimental results clearly showcase significant improvements in activity recognition performance for unseen classes across various IoT modalities. This advancement opens up new possibilities for intelligent and interactive IoT sensing with instruction from natural language.

\bibliographystyle{IEEEtran}
\bibliography{reference}

\end{document}